\documentclass[conference]{IEEEtran}

\usepackage{url}
\usepackage{caption}
\usepackage{subcaption}

\usepackage{graphics} %
\usepackage{epsfig} %

\usepackage[font=small]{caption}

\usepackage{times} %
\usepackage{xcolor} %
\usepackage{units} %
\usepackage{graphicx}
\usepackage{latexsym}
\usepackage{setspace}
\usepackage{multicol}
\usepackage{verbatim}
\usepackage{float}
\usepackage{cite}
\usepackage{subcaption}
\usepackage[export]{adjustbox}
\usepackage{soul}

\usepackage{algorithm} 
\usepackage{algcompatible}

\newcommand{\multiline}[1]{%
  \begin{tabularx}{\dimexpr\linewidth-\ALG@thistlm}[t]{@{}X@{}}
    #1
  \end{tabularx}
}
\usepackage{arydshln} %

\usepackage{amssymb,amsmath,bm}

\def\beq{\begin{equation}}
\def\eeq{\end{equation}}
\def\bql{\begin{equation}}
\def\eql{\end{equation}}
\def\bqn{\begin{eqnarray*}}
\def\eqn{\end{eqnarray*}}
\def\bnl{\begin{eqnarray}}
\def\enl{\end{eqnarray}}
\def\bna{\bql\begin{array}{rcl}}
\def\ena{\end{array}\eql}
\def\bnn{\beq\begin{array}{rcl}}
\def\enn{\end{array}\eeq}
\def\bma{\begin{bmatrix}}
\def\ema{\end{bmatrix}}
\def\bmx{\begin{matrix}}
\def\emx{\end{matrix}}
\def\ben{\begin{enumerate}}
\def\een{\end{enumerate}}
\def\bit{\begin{itemize}}
\def\eit{\end{itemize}}
\def\bei{\begin{itemize}}
\def\eei{\end{itemize}}
\def\bet{\begin{tabular}}
\def\eet{\end{tabular}}

\newcommand{\allcaps}[1]{\uppercase\expandafter{#1}}

\begin{document}

\title{Design of a Miniature Underwater Vehicle and Data Collection System for Indoor Experimentation}
\author{
\IEEEauthorblockN{Jacob Herbert and Artur Wolek}
\IEEEauthorblockA{Department of Mechanical Engineering and Engineering Science\\
University of North Carolina at Charlotte\\
Charlotte, NC 28223 USA \\
Email: jacobstoneherbert@gmail.com, awolek@charlotte.edu}
}

\maketitle

\begin{abstract}
This paper describes the design of a miniature uncrewed underwater vehicle (MiniUUV) and related instrumentation for indoor experimentation. The  MiniUUV was developed using 3D printed components and low-cost, off-the-shelf electronics. The vehicle uses a propeller differential propulsion drive and a peristaltic pump with a syringe for buoyancy control. A water tank with an overhead camera system was constructed to allow for convenient indoor data collection in a controlled environment. Several tests were conducted to demonstrate the capabilities of the MiniUUV and data collection system, including buoyancy pump actuation tests and straight line, circular, and zig-zag motion tests on the surface. During each planar motion test an AprilTag was attached to the MiniUUV and an overhead camera system obtained video recordings that were processed offline to estimate vehicle position, surge velocity, sway velocity, yaw angle, and yaw rate.
\end{abstract}

\section{Introduction}
Marine robots have many applications in science, security, and commerce, such as environmental monitoring, bathymetry mapping, underwater surveillance, subsea search, and infrastructure inspection \cite{Schranz,bogue2015underwater}.  Testing full-scale vehicles in outdoor environments is often costly and requires specialized infrastructure and equipment.  However, small miniaturized underwater vehicles  provide a compact and accessible platforms for research and education that requires minimal infrastructure for deployment and can enable operation in confined spaces, shallow water, and in swarm configurations. An indoor testbed with small marine robots can be used to conduct scaled experiments  in a controlled laboratory environment without interference from wind, currents, waves, and vegetation. 

 Numerous research groups have made use of indoor water tanks to support research in underwater vehicle design, communication, navigation, and control.  For example, Blue Swarm~\cite{BlueSwarm} is a robotic swarm that operates within an indoor water tank using agents that have light-emitting sensors and cameras to investigate three-dimensional collective behaviors. In~\cite{MicroUSV}, a miniature fleet of  unmanned surface vehicles was developed to investigate algorithms for marine surveillance and  environmental monitoring. In~\cite{Paley}, the authors built a fish-inspired soft robot swarm for testing  formation control behaviors.  In \cite{ScottKelly}, proportional heading control of a single degree-of-freedom fish-like robot  was studied. Other related work involving small-scale robots has focused on miniaturization \cite{hanff2017auv, duecker2020hippocampusx}, enabling swarms and multi-robot systems \cite{mintchev2014mechatronic,amory2016sembio}, bio-inspired designs \cite{song2016compact, iacoponi2022h, berlinger2017robust,wu2019towards},  underwater glider designs \cite{zhang2013miniature}, operation in confined spaces \cite{duecker2018micro}, using novel actuation mechanisms and platforms \cite{knizhnik2020design,griffiths2016avexis,jin2015six} and developing docking capabilities \cite{mintchev2015towards,knizhnik2021docking}. Prior work has also investigated  the integration of acoustic modems   \cite{tao2018evaluating} and  the use of flow-based control \cite{knizhnik2022flow} with small under-actuated marine robots.

\begin{figure}[h!]
    \centering
    \includegraphics[width=0.35\textwidth]{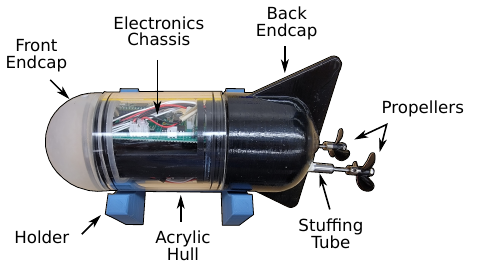}
    \caption{The MiniUUV and several of its components.}
    \label{fig:MiniUUV}
\end{figure}

This paper describes the development of a prototype miniature unmanned underwater vehicle (MiniUUV)  and indoor water tank for data collection that will support future research at UNC Charlotte in vehicle path planning and control, bio-inspired locomotion and sensing, and multi-vehicle coordination.  
The miniature underwater vehicle (see Fig.~\ref{fig:MiniUUV})  is capable of three-dimensional motion and is designed from low-cost and readily available components. The data processing method presented allows determining vehicle positions and velocities using a visual tracking system when the robot is on the surface. Results from experimental demonstrations illustrate the basic functionality of the  robot and data collection system.

The remainder of this paper is organized as follows. Section II discusses the design of the the MiniUUV, including all its subsystems and the data processing procedure. Section III presents the experimental data collected. Section IV concludes the paper and suggests future work.   

\section{Miniature Underwater Vehicle Design}
This section describes the design and fabrication of the MiniUUV and related instrumentation for data collection.
For a more comprehensive overview of the MiniUUV refer to  \cite{HerbertMSThesis}.

\subsection{MiniUUV Design Overview}

\subsubsection{Structural Design}
\begin{figure}[h!]
    \centering
    \includegraphics[width=0.45\textwidth]{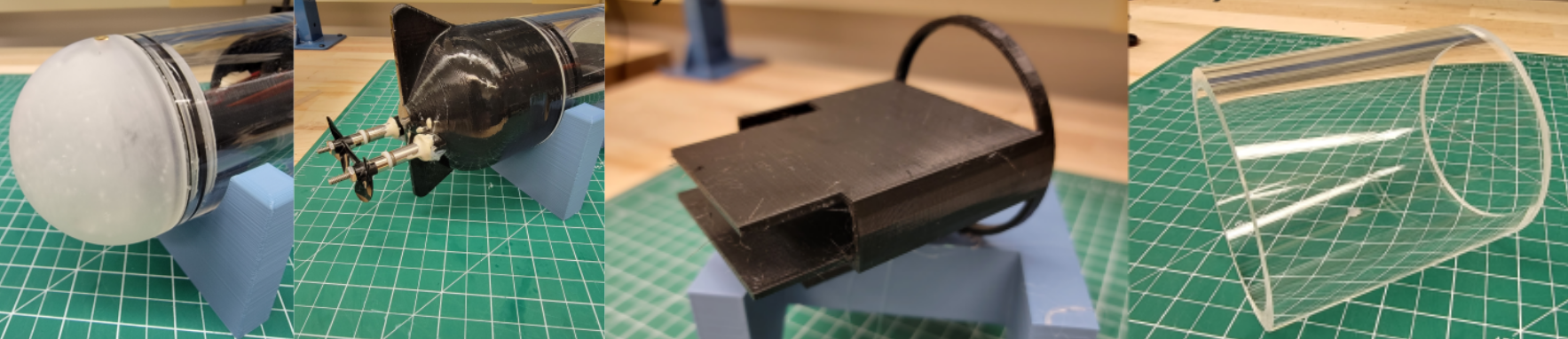}
    \caption{The front endcap (far left) is inserted into the hull (far right acrylic tube) and sealed using two 230 O-rings. The back endcap (middle left) houses the propulsion system sub-assembly along with the peristaltic pump used in the buoyancy system. The chassis (middle right) houses the main electronics suite, including, the radio, micro-controller, inertial measurement unit (IMU), battery, as well as the syringe and the reflectance sensor used in the buoyancy system. }
    \label{fig:StructuralDesign}
\end{figure}

The structure of the MiniUUV consists of a front and back endcap, an acrylic tube, and a frame/chassis to house electronics. The acrylic tube has a 3'' inner diameter and 1/4'' wall thickness (see Fig.~\ref{fig:StructuralDesign}). A clear tube was chosen to reduce attenuation of radio signals and allow visibility into the internal components. The tube is chamfered on the inner rim to allow for easier insertion of the O-rings.  The front and back endcaps and electronics chassis were all 3D printed using an Original Prusa i3 MK3S with PETG filament. This filament was chosen due to its high strength and relatively low cost. Since this manufacturing method creates a porous surface that can leak or absorb water, the exterior of the endcaps were coated with a two-part epoxy (West Systems 105 epoxy and West Systems 205 fast hardener) to ensure the vessel remains watertight.  The endcaps were designed with two O-ring grooves that accept lubricated number 230 O-rings to seal the endcap against the acrylic tube. The O-ring groove dimensions were sized according to the Parker O-ring Handbook \cite{ParkerO-Ring}. The back endcap houses the buoyancy and propulsion systems and contains a mixture of tungsten powder and epoxy as ballast. This ballast is used to achieve a desired weight distribution for the vehicle to be close to neutrally buoyant and sit level in the water.  The front endcap is a streamlined hemispherical shape. The electronics chassis houses various electronics among two stacked layers, including:  the main circuit board, a SX1278 433Mhz LoRa URAT Serial Transceiver radio, a 6V 2000mAh NiMh battery, a syringe, and a reflectance sensor (used for buoyancy control). To assemble the vehicle, the chassis is inserted into the acrylic tube and sandwiched between the front cap and back cap. 

\subsubsection{Buoyancy System}
The buoyancy system consists of a peristaltic pump, motor controllers, a syringe reservoir, and a reflectance sensor array. An Adafruit 6V peristaltic liquid pump was selected and is capable of pumping up to 100 mL/min. This pump was chosen due to its high flow rate and small size. The pump is  connected to the syringe reservoir via a silicon tube, and a 25 mL syringe acts as a water reservoir within the system. When the syringe is half full the MiniUUV is approximately neutrally buoyant. Adjusting the syringe position modulates the water intake and thereby the depth rate. A Pololu reflectance sensor array is used to monitor the position of the plunger in the syringe using 9 infrared (IR) sensors. Depth is measured by a Blue Robotics
Bar02 depth sensor. Currently the system is operated in an open-loop fashion. 
Future work aims to develop a closed-loop depth controller and utilize a more powerful pump to improve reliability at the deepest point in the water tank. 

\subsubsection{Propulsion System}
\begin{figure}[h!]
    \centering
    \includegraphics[scale=.55]{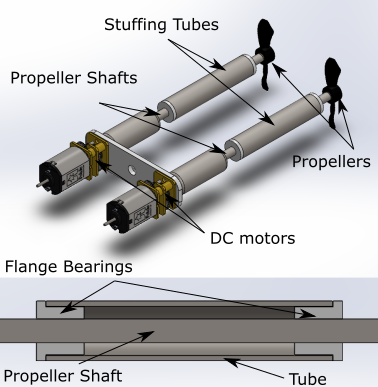}
    \caption{Top: The propulsion system uses differential drive propellers and two 6V DC motors. Bottom:  The stuffing tube houses a section of the propeller shaft along with a water resistant grease that prevents water from entering the vessel.}
    \label{fig:Propulsion}
\end{figure}

The propulsion system (adapted from~\cite{MicroUSV}) uses a differential propeller drive design for locomotion and includes two propeller shafts bound together by a 3D printer bracket, see Fig.~\ref{fig:Propulsion}. The propulsion system uses two 5:1 Micro Metal Gearmotor HP 6V DC motors to drive two custom-made propeller shafts. The shafts are made of 1/8'' stainless steel rods with a 6-32 thread on one end and are housed within stuffing tubes. The 1/4'' stainless steel stuffing tubes create a seal around the propeller shafts with water-resistant grease and contain two  steel flange bearings that are secured using blue-strength Loctite. 
\
\subsubsection{Electrical Design}
\begin{figure}[h!]
    \centering
    \includegraphics[scale=.50]{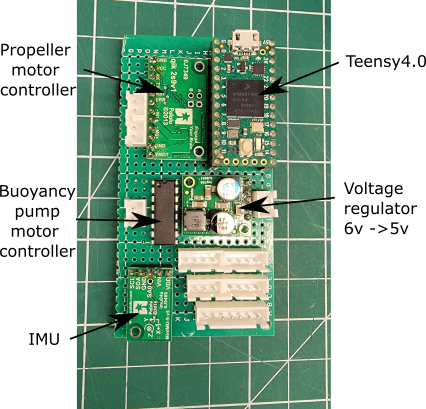}
    \caption{Main electronics board for the MiniUUV.}
    \label{fig:Motherboard}
\end{figure}

A Teensy 4.0 micro-controller controls actuators and processes all sensor data from the onboard inertial measurement unit  (Pololu MinIMU-9), the depth sensor, and the reflectance sensor array (Pololu QTR-MD-13RC). The micro-controller communicates sensor data and receives commands using the LoRa radio. Communication is only possible if the vehicle is on the surface or submerged a few feet in the water.  The propulsion motors are controlled by a Pololu Qik 2s9v1 Dual Serial motor controller whereas the peristaltic pump is controlled by a L293D 16-pin IC stepper motor driver. A majority of the electronics are mounted to a single breadboard as shown in Fig.~\ref{fig:Motherboard}.

\subsection{Water Tank and Overhead Tracking System}
\begin{figure}[h!]
    \centering
    \includegraphics[scale = .85]{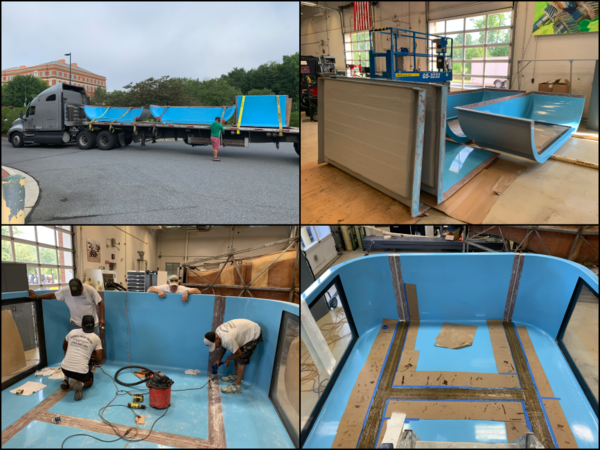}
    \caption{Water tank during delivery and assembly.}
    \label{fig:TankAssembly}
\end{figure}

A testing environment was developed for the MiniUUV that consisted of a 13.5 ft by 13.5 ft by 4.5 ft fiberglass water tank containing approximately 5000 gallons of water. The tank was manufactured by Dolphin Fiberglass and was assembled on-site by bolting together adjoining flanges of four sections, fastening the base to a plywood board, using a wet fiberglass layup to join and waterproof the seams, and further sanding and painting the seams (see Fig.~\ref{fig:TankAssembly}). Two large acrylic windows were installed on opposite sides of the tank.
 \begin{figure}[h!]
    \centering
    \includegraphics[scale = .35]{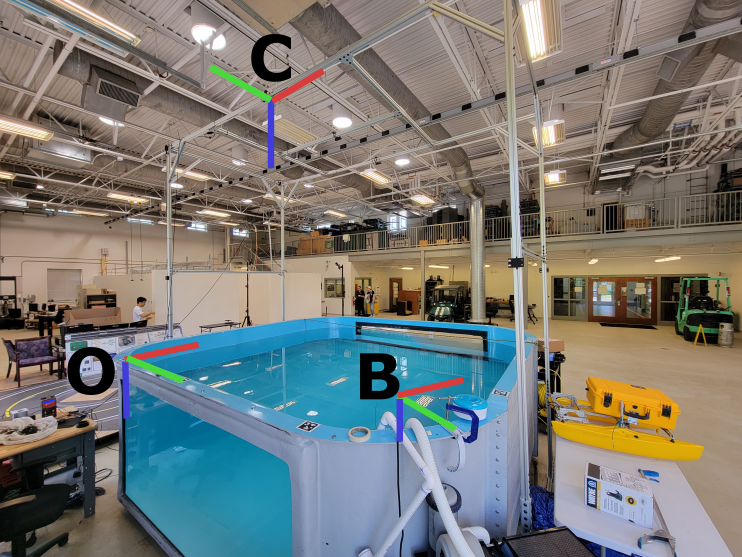}
    \caption{Locations and orientation of image frame $C$, world frame $O$, and body frame $B$.}
    \label{fig:TankFrames}
\end{figure}
A 15 ft tall frame made of aluminum 80-20 beams was built and placed over the top of the water tank. The frame was equipped with  three overhead Logitech webcams allowing complete and overlapping water tank coverage.
 The overhead cameras are connected to a central computer workstation located adjacent to the tank that runs MATLAB  to detect and localize an AprilTag\cite{AprilTag} attached to the MiniUUV  using the Image Processing and Computer Vision toolbox. The 36h11 family of AprilTags was chosen due to their robustness against false detection.  AprilTags are of a known size and shape and are easily detected and located by the MATLAB processing code and have been used in similar applications \cite{hanff2017auv}.  However, this system only functions when the MiniUUV is on the surface of the water. Future work may explore using AprilTags for underwater 3D positioning \cite{duecker2019towards}.
Each camera was calibrated using MATLAB's Camera Calibration application \cite{CameraCalibration} by taking 20 images of a checkerboard pattern in various orientations. The camera's extrinsic and intrinsic parameters were determined using these calibration images and subsequently used for data collection. A computer workstation adjacent to the water tank was used to record videos and to send/receive radio with the MiniUUV using a custom messaging protocol.

\subsection{Image Data Processing}
\label{eq:processing}
The video from a single camera was processed frame by frame to find the pose of the MiniUUV and all AprilTags within the camera's frame (other AprilTags were placed on the boundary of the tank as reference points). The pose of an AprilTag labeled $B$ at time $t$ consists of a translation vector ${\bm q}_{B/C}(t) = [x_{B/C}(t)~y_{B/C}(t)~z_{B/C}(t)]^{\rm T}$ from the camera frame to the center of the AprilTag, $B$, and a rotation matrix, ${\bm R}_{B/C}$ describing the rotation from the camera frame to the frame of AprilTag $B$. Both of these quantities are found using the \texttt{readAprilTag()} function in MATLAB \cite{readAprilTag}. This function uses the known size of the AprilTags (2.8 in) and intrinsic camera parameters to estimate each tag's  pose. 

\begin{figure}[h!]
    \centering
    \includegraphics[width=0.4\textwidth]{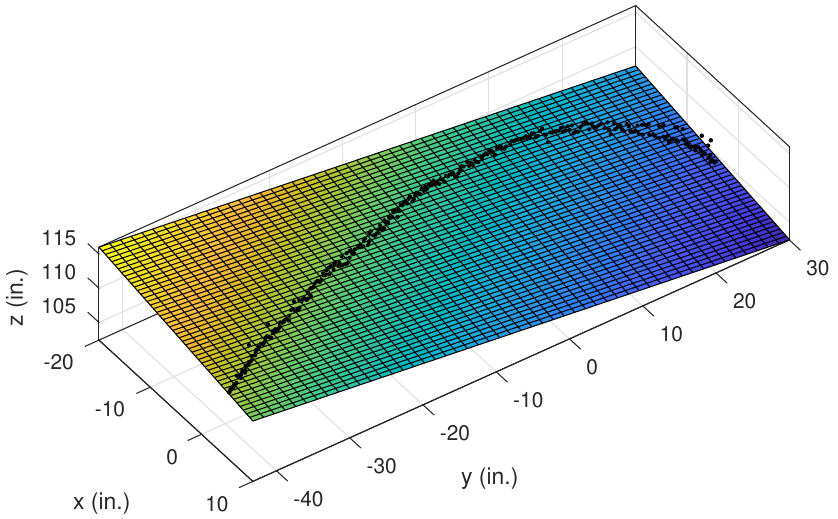}
    \caption{Example of plane fitting to position data.}
    \label{fig:PlaneFitting}
\end{figure}

The detected position of the MiniUUV possessed noise throughout each planar run and a slight tilt  in the vertical direction due to the mounting of the camera. This noise and tilt caused the motion of the robot to appear at varying $z$ values. To compensate for the tilt, a plane surface  of the form 
\begin{equation}
     a x+ b y + c z + d = 0 \;
\label{eq:plane}
\end{equation} 
was fit to the $(x_{B/C}, y_{B/C}, z_{B/C})$ data (see Fig.~\ref{fig:PlaneFitting}), where $a$, $b$, $c$, and $d$ are constant coefficients and $c = -1$ is fixed.

\begin{figure}[h!]
    \centering
    \includegraphics[scale=.5]{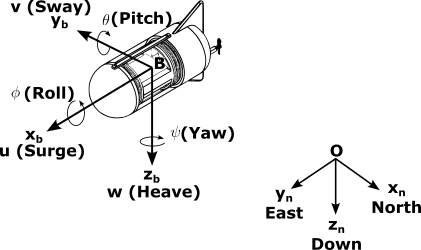}
    \caption{Left: MiniUUV body frame. Right: World frame.}
    \label{fig:Reference}
\end{figure}

Figure \ref{fig:Reference} shows the relationship between a world frame and the vehicle's body frame. This figure uses the north-east-down (NED) frame as the world frame with an origin labeled as $O$ and axes $x_{n}$, $y_{n}$, and $z_{n}$. The body-fixed frame is described by the translating origin, $B$, and rotating axes $x_b$, $y_b$, and $z_b$. The velocity along the $x_b$-axis is referred to as the surge velocity, $u$, the velocity along the $y_b$-axis is referred to as the sway velocity, $v$, and the velocity in the $z_b$-direction is referred to as the heave velocity, $w$. The rotation about the $x_b$-axis is denoted as the roll angle, $\phi$, the rotation about the $y_b$-axis is denoted as the pitch angle, $\theta$, and the rotation about the $z_b$-axis is denoted as the heading or yaw angle, $\psi$, respectively. The world frame origin can potentially be chosen as a static point anywhere in the water tank testing area. For simplicity, we set the origin as the position of the vehicle in the first image frame, $x_{O/C} = x_{B/C}(0)$, $y_{O/C} =  y_{B/C}(0)$, and $z_{O/C} =  z_{B/C}(0)$.
The data was then transformed from the camera's frame and origin to the world  frame and origin according to: 
\begin{equation}
    \begin{bmatrix}
        x_{B/O}(t) \\
        y_{B/O}(t) \\
        z_{B/O}(t) \\
    \end{bmatrix}
    = \bm R_{O/C}^{\rm T} 
\left(
    \begin{bmatrix}
        x_{B/C} (t)\\
        y_{B/C} (t)\\
        z_{B/C} (t)\\
    \end{bmatrix}
-
    \begin{bmatrix}
        x_{O/C}  \\
        y_{O/C} \\
        z_{O/C} \\
    \end{bmatrix}
\right)
\;,
\end{equation}
where $\bm R_{O/C}$ is the rotation matrix relating the world frame to the camera frame. This matrix is computed according to 
\begin{align}
   \bm R_{O/C} &= 
    \begin{bmatrix}
        \bm u_1^{\rm T}\\
        \bm u_2^{\rm T}\\
        \bm u_3^{\rm T}\\
    \end{bmatrix} \;,
~\text{where}~
    \bm u_3 = \dfrac{
    \begin{bmatrix}
        a & b& c 
    \end{bmatrix}^T
    }{||
    \begin{bmatrix}
        a& b& c
    \end{bmatrix}
    ||} 
\label{eq:rot_mat}
\end{align}
is the normal vector to the plane using coefficients from \eqref{eq:plane} and the remaining vectors are defined by 
$
    \bm u_1 = 
[
         a ~ b ~ c 
]^{\rm T}
    \times
[
        1 ~ 0 ~ 0 
]^{\rm T}
$
and
$    \bm u_2 = 
    \bm u_1
    \times
    \bm u_3
$. Note that additional rotations may be applied to \eqref{eq:rot_mat} to align the axes in a preferred convention relative to the testing area. 
The heading angle $\psi$ was extracted directly from ${\bm R}_{B/C}$ without further corrections. 

 The data $x_{B/O}$, $y_{B/O}$ and $\psi$ were differentiated using a finite difference approximation  and then passed through a moving average smoothing filter with a 12 point window size to give $\dot x_{B/O}$, $\dot y_{B/O}$, and $r = \dot \psi$.  
Lastly, the body frame surge and sway velocities were obtained from 
\begin{equation}
    \begin{bmatrix}
        u \\
        v \\
        w \\
    \end{bmatrix} 
=
    \begin{bmatrix}
        \cos(\psi) & \sin(\psi) & 0 \\
        -\sin(\psi) & \cos(\psi) & 0 \\
        0 & 0 & 1 \\
    \end{bmatrix}
    \begin{bmatrix}
        {\dot x}_{B/O} \\
        {\dot y}_{B/O} \\
        0
    \end{bmatrix}\;.
\end{equation}
Due to small variations in the frame rate of the camera, the data was re-sampled to 30 Hz to produce a consistent output. The data processing method presented above is only applicable if the MiniUUV is operating on the surface; if the MiniUUV is submerged the distortion from the water and variations  in depth must be accounted for (not considered in this work).
\section{Indoor Demonstrations}
This section describes several open-loop planar motion and varying buoyancy tests  that demonstrates the basic functionality of the MiniUUV and data collection system.
\subsection{Planar Motion Tests}
\begin{figure}[H]
    \centering
    \includegraphics[width=0.35\textwidth]{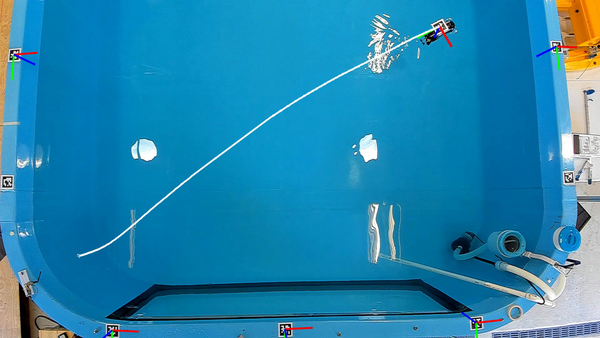}
 \includegraphics[width=0.35\textwidth]{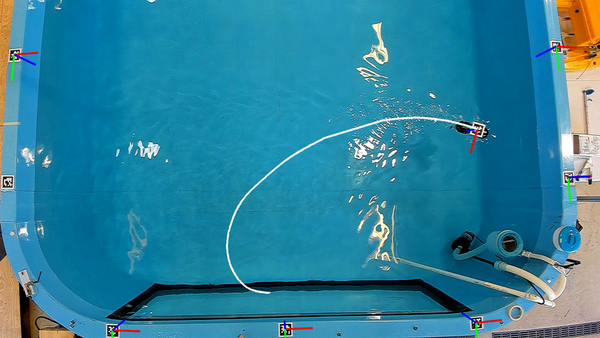}
    \includegraphics[width=0.35\textwidth]{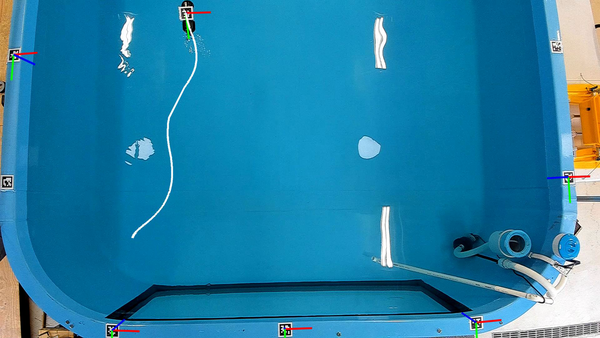}
    \caption{The top, middle, and bottom correspond to the final images recorded from three trials where the MicroUUV executed a straight line path, circular arc, and zig-zag pattern, respectively.}
    \label{fig:Line_Real}
\end{figure}

\begin{figure}[h!]
\centering
    \includegraphics[scale = 0.52]{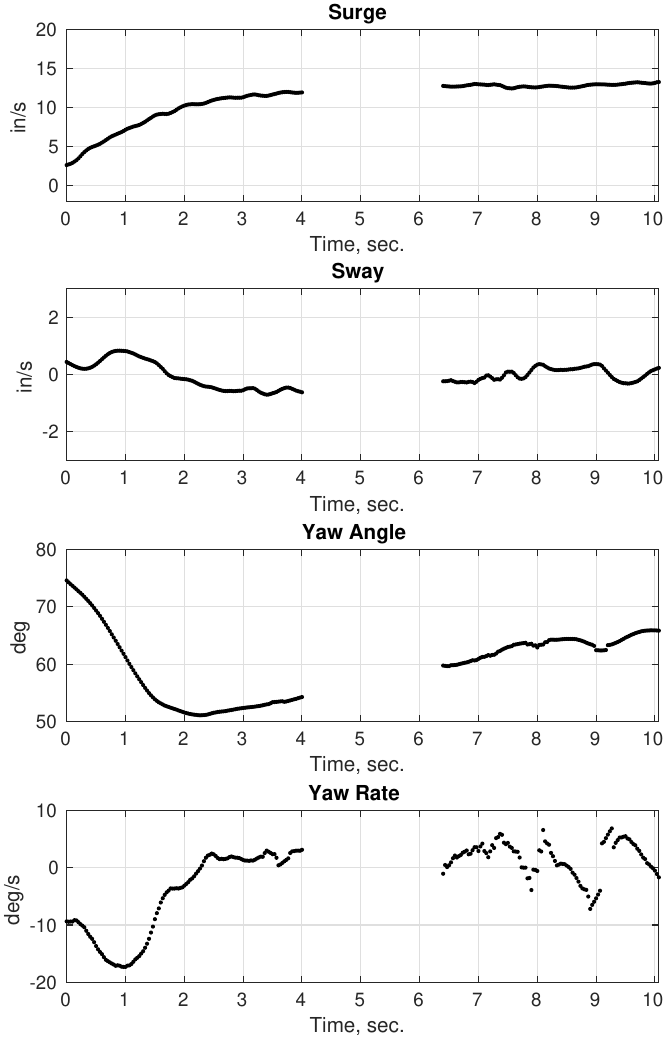}
    \caption{Data collected from the linear planar motion test.}
    \label{fig:Case_1}
\end{figure}

\begin{figure}[h!]
    \centering
    \includegraphics[scale = 0.52]{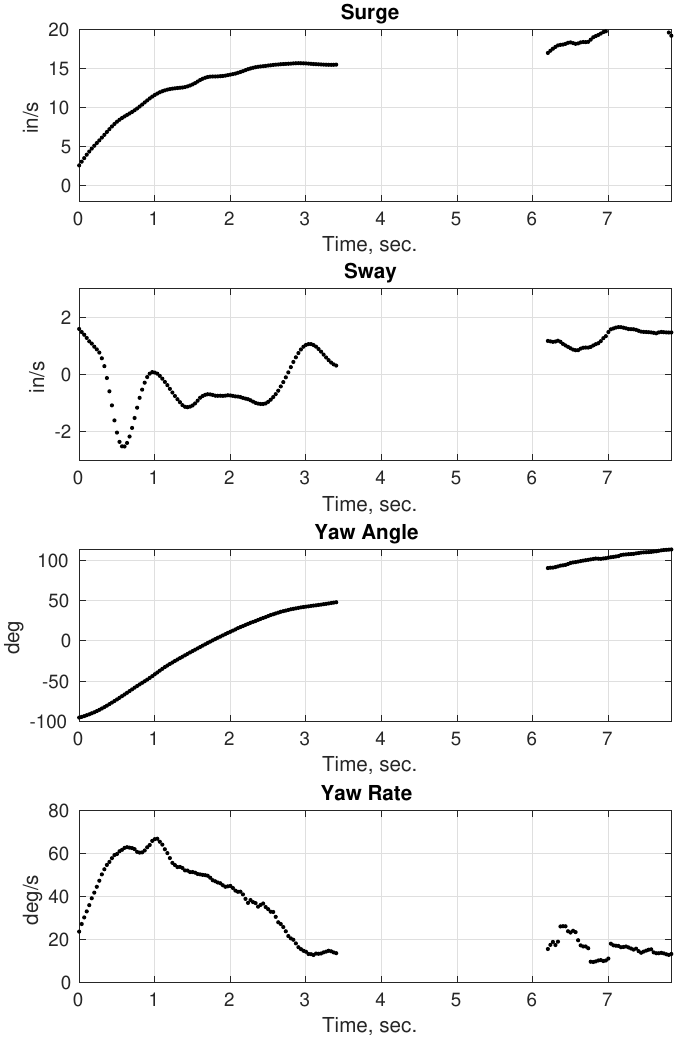}
    \caption{Data collected from the circular planar motion test.}
    \label{fig:Case_2}
\end{figure}

\begin{figure}[h!]
    \centering
    \includegraphics[scale = 0.52]{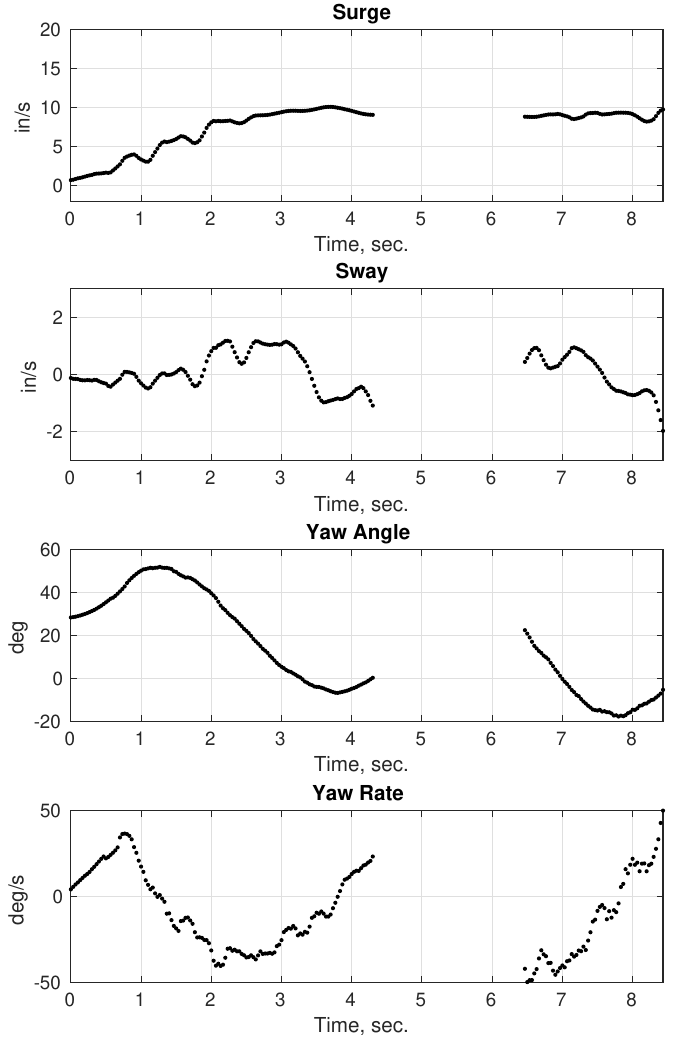}
    \caption{Data collected from the zig-zag planar motion test.}
    \label{fig:Case_3}
\end{figure}

Examples of straight line, circular, and zig-zag paths are shown in Fig.~\ref{fig:Line_Real}. The demonstrations were performed using a pre-programmed sequence of motor commands.  In Fig.~\ref{fig:Line_Real} the AprilTag mounted on the moving MiniUUV can be seen along with several other static AprilTags along the  tank perimeter.
Starting from the same edge of the tank the MiniUUV's actuation commenced once a radio signal was received from the ground station. The linear path was approximately 10 ft long, the circular path had a radius of about 6 ft, and the zig-zag path contained two zig-zag maneuvers. The AprilTag data collected was processed as described in Sec.~\ref{eq:processing} and the resulting survey velocity, sway velocity, yaw angle, and yaw rate are shown in Figs.~\ref{fig:Case_1}--\ref{fig:Case_3}. Note that the yaw angle is measured positive clockwise from the (approximately) up direction when viewing  Figs.~\ref{fig:Case_1}--\ref{fig:Case_3}.
Such data can be useful for system identification. Each run contained sections of data where AprilTag positions were either spurious or absent due to glare from overhead lighting or wave reflections interfering with the detection of the AprilTag. To remove the artifacts this caused on the smoothed velocity outputs some of the data was manually trimmed from the plots shown below. 
\subsection{Buoyancy Tests}
\begin{figure}[h!]
    \centering
\includegraphics[width=0.35\textwidth]{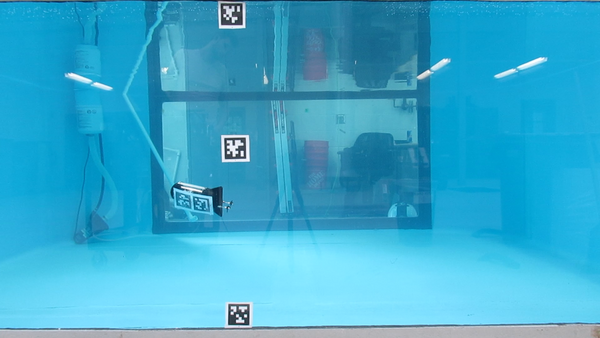}\\
    \caption{MiniUUV during pump actuation tests.}
    \label{fig:depth_test}
\end{figure}

\begin{figure}[h!]
    \centering
    \includegraphics[width = 0.45\textwidth]{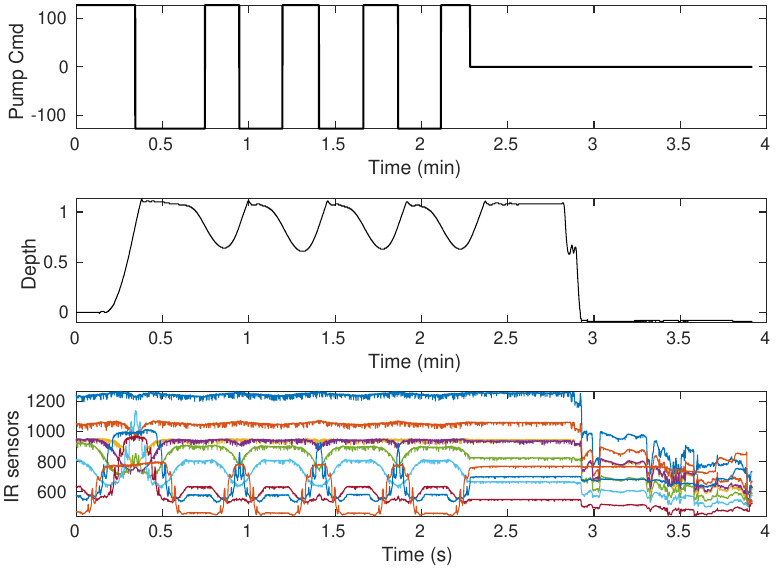}
    \caption{Data recorded during MiniUUV pump actuation tests.}
    \label{fig:depth_test_results}
\end{figure}

Open-loop pump actuation tests were also conducted to test the buoyancy system. In these tests, the MiniUUV descended to the bottom of the water tank and then ascended towards the surface several times while recording depth and infrared sensor readings of the syringe plunger position. The experimental setup is shown in Fig.~\ref{fig:depth_test} and the recorded data are shown in Fig.~\ref{fig:depth_test_results}. 
The data shows the MiniUUV descending from the surface to approximately 1 meter depth and periodically rising/sinking as the syringe plunger moves in and out based on pump commands. The position of the syringe is detected by the 9-channel IR sensor data shown on the bottom panel of Fig.~\ref{fig:depth_test_results}. The results suggest that a calibration table could be developed to predict plunger position from the nine IR inputs. However, once the vehicle surfaces near the end of the trial the bobbing of the vehicle on the surface causes changing lighting conditions that affects all IR sensors---an opaque housing may alleviate this issue in future designs.

\section{Conclusion}
An indoor water tank facility and a miniature underwater vehicle testbed were developed for future research in vehicle path planning and control, bio-inspired locomotion and sensing, and multi-vehicle coordination.  The MiniUUV was constructed from low-cost off-the-shelf electronics and 3D-printed parts. It uses a differential propeller  drive, a peristaltic pump with a syringe and infrared sensors for buoyancy control, a Teensy onboard micro-controller, and a 433 MHz radio for communication. The data collection system processes videos to detect an AprilTag and produce vehicle position, velocity, and orientation estimates when the MiniUUV is on the water surface.  The basic functionality of the MiniUUV was demonstrated through several open-loop  planar motion experiments and buoyancy pump actuation tests. The MiniUUV has served as a prototype to demonstrate feasibility  of some design elements and to highlight areas that could be improved---including the need for a more robust pump and syringe plunger position sensor and a lower frequency and/or higher output power radio. The design presented and lessons learned may be useful to researchers developing similar systems.

Future work may include developing follow-on MiniUUV variants that incorporate various design improvements, are more compact and maneuverable, and are capable of being deployed in multi-vehicle configurations.
\section{Acknowledgments}
This work was supported in part by UNC Charlotte's Faculty Research Grants Program. We thank Reggie Hart for his logistical support and design of a tank filtration system.
\bibliographystyle{IEEEtran}
\bibliography{main}

\begin{thebibliography}{10}
\providecommand{\url}[1]{#1}
\csname url@samestyle\endcsname
\providecommand{\newblock}{\relax}
\providecommand{\bibinfo}[2]{#2}
\providecommand{\BIBentrySTDinterwordspacing}{\spaceskip=0pt\relax}
\providecommand{\BIBentryALTinterwordstretchfactor}{4}
\providecommand{\BIBentryALTinterwordspacing}{\spaceskip=\fontdimen2\font plus
\BIBentryALTinterwordstretchfactor\fontdimen3\font minus
  \fontdimen4\font\relax}
\providecommand{\BIBforeignlanguage}[2]{{%
\expandafter\ifx\csname l@#1\endcsname\relax
\typeout{** WARNING: IEEEtran.bst: No hyphenation pattern has been}%
\typeout{** loaded for the language `#1'. Using the pattern for}%
\typeout{** the default language instead.}%
\else
\language=\csname l@#1\endcsname
\fi
#2}}
\providecommand{\BIBdecl}{\relax}
\BIBdecl

\bibitem{Schranz}
M.~Schranz, M.~Umlauft, M.~Sende, and W.~Elmenreich, ``Swarm robotic behaviors
  and current applications,'' \emph{Frontiers in Robot. and AI}, vol.~7, pp.
  1--20, 2020.

\bibitem{bogue2015underwater}
R.~Bogue, ``Underwater robots: a review of technologies and applications,''
  \emph{Industrial Robot: An Int. J.}, vol.~42, no.~3, pp. 186--191, 2015.

\bibitem{BlueSwarm}
F.~Berlinger, M.~Gauci, and R.~Nagpal, ``Implicit coordination for {{3D}}
  underwater collective behaviors in a fish-inspired robot swarm,'' vol.~6,
  no.~50.\hskip 1em plus 0.5em minus 0.4em\relax Science Robot., 2021, p.
  eabd8668.

\bibitem{MicroUSV}
C.~Gregory and A.~Vardy, ``{microUSV}: A low-cost platform for indoor marine
  swarm robot. research,'' \emph{HardwareX}, vol.~7, p. e00105, 2020.

\bibitem{Paley}
D.~A. Paley, A.~A. Thompson, A.~Wolek, and P.~Ghanem, ``Planar formation
  control of a school of robotic fish: Theory and experiments,''
  \emph{Frontiers in Control Eng.}, vol.~2, p.~15, 2021.

\bibitem{ScottKelly}
S.~D. Kelly, M.~J. Fairchild, P.~M. Hassing, and P.~Tallapragada,
  ``Proportional heading control for planar navigation: The {Chaplygin} beanie
  and fishlike robotic swimming,'' in \emph{2012 Amer. Control Conf.}, 2012,
  pp. 4885--4890.

\bibitem{hanff2017auv}
H.~Hanff, P.~Kloss, B.~Wehbe, P.~Kampmann, S.~Kroffke, A.~Sander, M.~B.
  Firvida, M.~von Einem, J.~F. Bode, and F.~Kirchner, ``{AUV$^{\rm X}$}—a
  novel miniaturized autonomous underwater vehicle,'' in \emph{MTS/IEEE OCEANS
  2017, Aberdeen}, 2017, pp. 1--10.

\bibitem{duecker2020hippocampusx}
D.~A. Duecker, N.~Bauschmann, T.~Hansen, E.~Kreuzer, and R.~Seifried,
  ``{HippoCampusX}--a hydrobatic open-source micro {AUV} for confined
  environments,'' in \emph{2020 IEEE/OES Auton. Underwater Vehicles Symp.},
  2020, pp. 1--6.

\bibitem{mintchev2014mechatronic}
S.~Mintchev, E.~Donati, S.~Marrazza, and C.~Stefanini, ``Mechatronic design of
  a miniature underwater robot for swarm operations,'' in \emph{2014 IEEE Int.
  Conf. on Robot. and Automat.}, 2014, pp. 2938--2943.

\bibitem{amory2016sembio}
A.~Amory and E.~Maehle, ``{SEMBIO}-a small energy-efficient swarm {AUV},'' in
  \emph{MTS/IEEE OCEANS 2016, Monterey}, 2016, pp. 1--7.

\bibitem{song2016compact}
Z.~Song, C.~Mazzola, E.~Schwartz, R.~Chen, J.~Finlaw, M.~Krieg, and K.~Mohseni,
  ``A compact autonomous underwater vehicle with cephalopod-inspired
  propulsion,'' \emph{Marine Technol. Society J.}, vol.~50, no.~5, pp. 88--101,
  2016.

\bibitem{iacoponi2022h}
S.~Iacoponi, G.~J. Van~Vuuren, G.~Santaera, N.~Mankovskii, I.~Zhilin, F.~Renda,
  C.~Stefanini, and G.~De~Masi, ``H-surf: Heterogeneous swarm of underwater
  robotic fish,'' in \emph{MTS/IEEE OCEANS 2022, Hampton Roads}, 2022, pp.
  1--5.

\bibitem{berlinger2017robust}
F.~Berlinger, J.~Dusek, M.~Gauci, and R.~Nagpal, ``Robust maneuverability of a
  miniature, low-cost underwater robot using multiple fin actuation,''
  \emph{IEEE Robot. and Automat. Lett.}, vol.~3, no.~1, pp. 140--147, 2017.

\bibitem{wu2019towards}
Z.~Wu, J.~Yu, J.~Yuan, and M.~Tan, ``Towards a gliding robotic dolphin: Design,
  modeling, and experiments,'' \emph{IEEE/ASME Trans. on Mechatronics},
  vol.~24, no.~1, pp. 260--270, 2019.

\bibitem{zhang2013miniature}
F.~Zhang, J.~Thon, C.~Thon, and X.~Tan, ``Miniature underwater glider: Design
  and experimental results,'' \emph{IEEE/ASME Trans. on Mechatronics}, vol.~19,
  no.~1, pp. 394--399, 2013.

\bibitem{duecker2018micro}
D.~A. Duecker, A.~Hackbarth, T.~Johannink, E.~Kreuzer, and E.~Solowjow, ``Micro
  underwater vehicle hydrobatics: A submerged furuta pendulum,'' in \emph{2018
  IEEE Int. Conf. on Robot. and Automat.}, 2018, pp. 7498--7503.

\bibitem{knizhnik2020design}
G.~Knizhnik and M.~Yim, ``Design and experiments with a low-cost single-motor
  modular aquatic robot,'' in \emph{2020 17th Int. Conf. on Ubiquitous Robots},
  2020, pp. 233--240.

\bibitem{griffiths2016avexis}
A.~Griffiths, A.~Dikarev, P.~R. Green, B.~Lennox, X.~Poteau, and S.~Watson,
  ``{AVEXIS}—{A}qua vehicle explorer for in-situ sensing,'' \emph{IEEE Robot.
  and Automat. Lett.}, vol.~1, no.~1, pp. 282--287, 2016.

\bibitem{jin2015six}
S.~Jin, J.~Kim, J.~Kim, and T.~Seo, ``Six-degree-of-freedom hovering control of
  an underwater robotic platform with four tilting thrusters via selective
  switching control,'' \emph{IEEE/ASME Trans. on mechatronics}, vol.~20, no.~5,
  pp. 2370--2378, 2015.

\bibitem{mintchev2015towards}
S.~Mintchev, R.~Ranzani, F.~Fabiani, and C.~Stefanini, ``Towards docking for
  small scale underwater robots,'' \emph{Auton. Robots}, vol.~38, pp. 283--299,
  2015.

\bibitem{knizhnik2021docking}
G.~Knizhnik and M.~Yim, ``Docking and undocking a modular underactuated
  oscillating swimming robot,'' in \emph{2021 IEEE Int. Conf. on Robot. and
  Automat.}, 2021, pp. 6754--6760.

\bibitem{tao2018evaluating}
Q.-y. Tao, Y.-h. Zhou, F.~Tong, A.-j. Song, and F.~Zhang, ``Evaluating
  acousticcommunication performance of micro autonomous underwater vehicles in
  confined spaces,'' \emph{Frontiers of Information Technol. \& Electronic
  Eng.}, vol.~19, no.~8, pp. 1013--1023, 2018.

\bibitem{knizhnik2022flow}
G.~Knizhnik, P.~Li, X.~Yu, and M.~A. Hsieh, ``Flow-based control of marine
  robots in gyre-like environments,'' in \emph{2022 Int. Conf. on Robot. and
  Automat.}, 2022, pp. 3047--3053.

\bibitem{HerbertMSThesis}
J.~Herbert, ``Design and system identification of a miniature underwater
  vehicle for controls research,'' Master's thesis, UNC Charlotte, 2023.

\bibitem{ParkerO-Ring}
\BIBentryALTinterwordspacing
Parker, ``Parker o-ring handbook,'' 2022, last accessed 10 July 2022. [Online].
  Available:
  \url{https://www.parker.com/Literature/O-Ring%20Division%20Literature/ORD%205700.pdf}
\BIBentrySTDinterwordspacing

\bibitem{AprilTag}
E.~Olson, ``Apriltag,'' https://april.eecs.umich.edu/software/apriltag, 2010,
  date Accessed: 2023-04-19.

\bibitem{duecker2019towards}
D.~A. Duecker, K.~Eusemann, and E.~Kreuzer, ``Towards an open-source micro
  robot oceanarium: A low-cost, modular, and mobile underwater motion-capture
  system,'' in \emph{2019 IEEE/RSJ Int. Conf. on Intell. Robots and Syst.},
  2019, pp. 8048--8053.

\bibitem{CameraCalibration}
{The MathWorks Inc.}, ``Camera calibration,''
  https://www.mathworks.com/help/vision/camera-calibration.html, 2023, date
  Accessed: 2023-05-11.

\bibitem{readAprilTag}
------, ``readapriltag,'' www.mathworks.com/help/vision/ref/readapriltag.html,
  2023, date Accessed: 2023-05-11.

\end{thebibliography}
\end{document}